%% file: main.tex
\documentclass{article}

\usepackage{arxiv}

\usepackage[utf8]{inputenc} 
\usepackage[T1]{fontenc}    
\usepackage[dvipsnames]{xcolor}
\definecolor{linkColor}{rgb}{0.18,0.39,0.62}
\usepackage[most]{tcolorbox}
\usepackage[colorlinks=true,linkcolor=linkColor,citecolor=linkColor,filecolor=linkColor,urlcolor=linkColor]{hyperref}
\usepackage{float}
\usepackage{url}            
\usepackage{booktabs}       
\usepackage{amsfonts}       
\usepackage{nicefrac}       
\usepackage{microtype}      
\usepackage{cite}
\usepackage{url}    
\usepackage{xcolor}
\usepackage{wrapfig}
\usepackage{color}
\usepackage{times}
\usepackage{colortbl}
\usepackage{amsmath}
\usepackage{tabularx}
\usepackage{makecell}
\usepackage{latexsym}
\usepackage{indentfirst}
\usepackage{enumerate}
\usepackage{multirow}
\usepackage{graphicx}
\usepackage{natbib}
\usepackage{booktabs}
\usepackage{color}
\usepackage{listings}
\usepackage{amssymb}
\usepackage{inconsolata}
\usepackage{ulem} 
\usepackage{microtype}
\usepackage{textcomp}  
\usepackage{scalerel}  

\usepackage{microtype}
\usepackage{hyperref}
\usepackage{url}
\usepackage{booktabs}
\usepackage{graphicx}
\usepackage{cancel}
\usepackage{xcolor} 
\usepackage{colortbl} 
\usepackage{amsmath}
\usepackage{amssymb}
\usepackage{mathtools}
\usepackage{amsthm}
\usepackage{enumitem}
\usepackage{pifont}
\usepackage{listings}
\usepackage{fontawesome}

\usepackage{wrapfig}
\usepackage{caption}
\usepackage{mdframed}
\usepackage{amssymb}

\usepackage{subcaption}
\usepackage{multirow}
\usepackage{multicol}
\usepackage{xspace}

\usepackage{subcaption}
\usepackage{multirow}
\usepackage{multicol}
\usepackage{xspace}
\usepackage{tcolorbox}
\usepackage[export]{adjustbox}
\usepackage{wrapfig,lipsum,booktabs}

\tcbset{
  aibox/.style={
    width=\linewidth,
    top=8pt,
    bottom=4pt,
    colback=blue!6!white,
    colframe=black,
    colbacktitle=black,
    enhanced,
    center,
    attach boxed title to top left={yshift=-0.1in,xshift=0.15in},
    boxed title style={boxrule=0pt,colframe=white,},
  }
}
\newcolumntype{C}[1]{>{\centering\let\newline\\\arraybackslash\hspace{0pt}}m{#1}}
\newtcolorbox{AIbox}[2][]{aibox,title=#2,#1}

\input{config}

\useunder{\uline}{\ul}{}

\definecolor{mypurple}{RGB}{128,0,128}

\title{ \textbf{\textcolor{mypurple}{G1}}: Bootstrapping Perception and Reasoning Abilities of Vision-Language Model via Reinforcement Learning}

\author{Liang Chen\textsuperscript{1,3}$^{*\dagger}$ \quad Hongcheng Gao\textsuperscript{2,3}$^*$\quad Tianyu Liu\textsuperscript{1}\quad Zhiqi Huang\textsuperscript{3}\quad \\    \textbf{Flood Sung\textsuperscript{3}\quad Xinyu Zhou\textsuperscript{3}\quad  Yuxin Wu\textsuperscript{3} \quad Baobao Chang\textsuperscript{1} } \\
  \textsuperscript{1}Peking University
  \textsuperscript{2}UCAS
  \textsuperscript{3}Moonshot AI 
}

\renewcommand{\headeright}
{\textbf{\textcolor{mypurple}{G1}}}

\renewcommand{\undertitle}{}
\renewcommand{\shorttitle}
  
\begin{document}

\maketitle

\def\thefootnote{$^*$}\footnotetext{Equal contribution. $^\dagger$Work done during internship at Moonshot AI. }
\def\thefootnote{\arabic{footnote}}

\input{files/abstract.tex}

\input{files/intro.tex}

\input{files/Method}

\input{files/exp}

\input{files/related_work.tex}

\input{files/conclusion}

{
\small
\normalem
\bibliographystyle{plain}
\bibliography{custom}

}
\input{appendix}

\newpage

\end{document}

%% file: config.tex
\usepackage{amsmath,amsfonts,bm}

\def\eqref#1{equation~\ref{#1}}

\def\1{\bm{1}}

\def\rvo{{\mathbf{o}}}

\def\rvq{{\mathbf{q}}}

\DeclareMathAlphabet{\mathsfit}{\encodingdefault}{\sfdefault}{m}{sl}
\SetMathAlphabet{\mathsfit}{bold}{\encodingdefault}{\sfdefault}{bx}{n}

\def\gQ{{\mathcal{Q}}}

\newcommand{\E}{\mathbb{E}}



%% file: files/abstract.tex
\begin{abstract}

Vision-Language Models (VLMs) excel in many direct multimodal tasks but struggle to translate this prowess into effective decision-making within interactive, visually rich environments like games. This ``knowing-doing'' gap significantly limits their potential as autonomous agents, as leading VLMs often performing badly in simple games.  To address this, we introduce VLM-Gym, a curated reinforcement learning (RL) environment featuring diverse visual games with unified interfaces and adjustable, compositional difficulty, specifically designed for scalable multi-game parallel training. Leveraging VLM-Gym, we train G0 models using pure RL-driven self-evolution, which demonstrate emergent perception and reasoning patterns. To further mitigate challenges arising from game diversity, we develop G1 models. G1 incorporates a perception-enhanced cold start prior to RL fine-tuning. Our resulting G1 models consistently surpass their teacher across all games and outperform leading proprietary models like Claude-3.7-Sonnet-Thinking. Systematic analysis reveals an intriguing finding: perception and reasoning abilities mutually bootstrap each other throughout the RL training process. Source code including VLM-Gym and RL training are released at \href{https://github.com/chenllliang/G1}{chenllliang/G1} to foster future research in advancing VLMs as capable interactive agents.

\end{abstract}

\begin{figure}[h]
\centering
\includegraphics[width=1\textwidth]{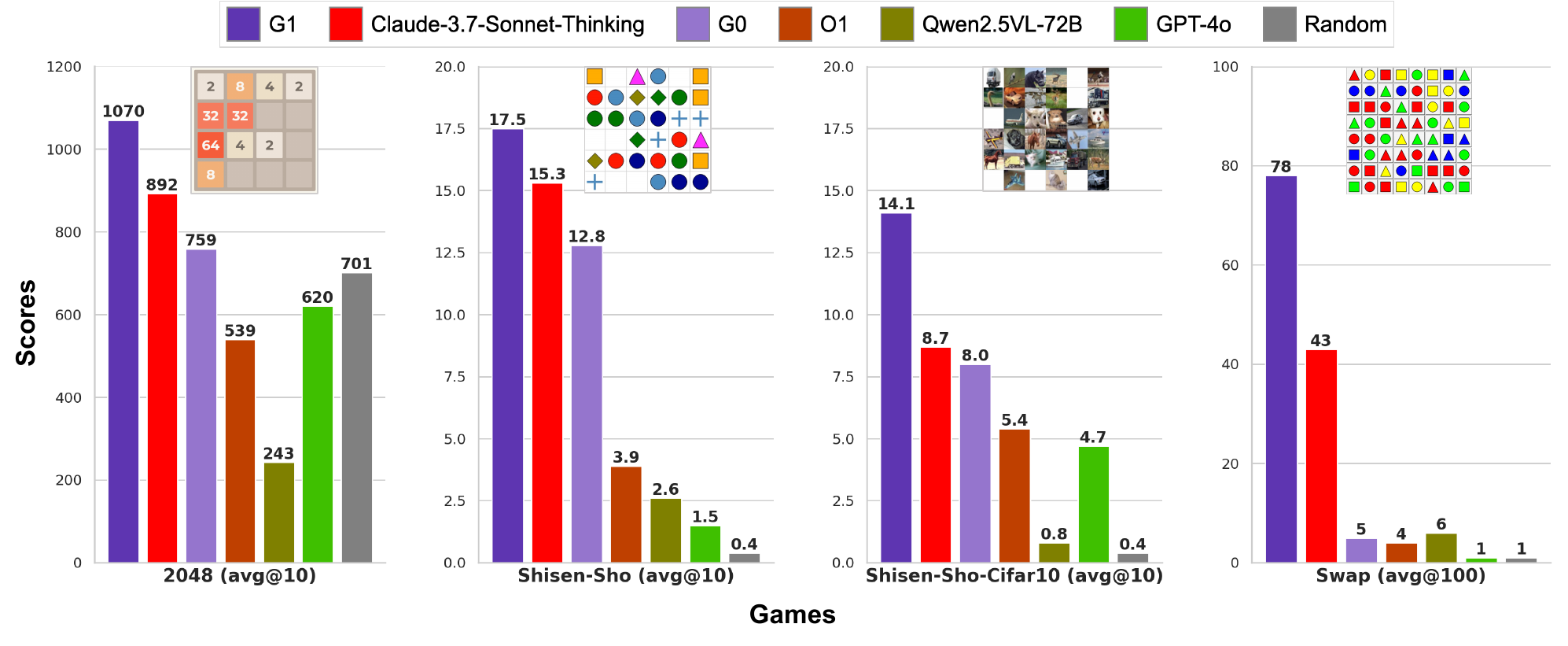}
\caption{Comparison of different models on games from VLM-Gym. }
\label{fig:teaser}
\end{figure}

%% file: files/intro.tex
\section{Introduction}

Interactive environments, particularly visually rich games ranging from classic Atari~\citep{mnih2013playingatarideepreinforcement} platforms and board games like Go (famously tackled by AlphaGo~\citep{Silver2016MasteringTG}) to recently popular complex simulations like Pokémon~\citep{Claude3S}, have been serving as popular testbeds for evaluating and advancing the general decision making capabilities of Artificial Intelligence (AI) agents, spanning visual perception, reasoning, and action. While Vision-Language Models (VLMs)~\citep{gpt4v,Claude3S,Qwen2.5-VL,geminiteam2024gemini,kimiteam2025kimivltechnicalreport,chen2024tokenpredictionmultimodalintelligence}, powered by advances in Large Language Models (LLMs), have demonstrated remarkable success in tasks like image captioning, visual question answering, and mathematical reasoning, translating this prowess to effective decision-making within interactive environments such as embodied robot~\citep{chen2024pcabench} and visual games remains a significant hurdle, limiting the boarder applications of VLMs in developing truly autonomous agents capable of reasoning and acting effectively based on rich visual information in the real world. In fact, VLM performance in such settings often falls considerably short of human capabilities~\citep{wang2025largevisionlanguagemodels,waytowich2024atarigptbenchmarkingmultimodallarge,paglieri2024balrog}. Our findings also underscore this challenge: even leading VLMs such as OpenAI-o1~\citep{openai2024openaio1card} and Qwen2.5VL-72B~\citep{Qwen2.5-VL} can struggle profoundly with straightforward games like Shisen-Sho and 2048, sometimes even achieving scores lower than random choice. This significant performance gap raises our interest in its potential causes and in methods for improving performance.

Perception and reasoning are two core abilities of VLMs. However, solving visual games requires not only accurate visual perception and reasoning about the current state and game rules, but also the crucial ability to translate that understanding into effective action, bridging the ``knowing-doing'' gap. While manually curating multimodal chain-of-thought data is one approach for training game-playing agents~\citep{chen2024pcabench, zhai2024finetuning}, this method faces scalability limitations. We therefore consider Reinforcement Learning (RL) with both historical and current relevance. RL is a well-established technique in game AI training (e.g., CNN-based Atari players~\citep{mnih2013playingatarideepreinforcement}). It is also gaining significant recent attention through methods such as Reinforcement Learning with Verifiable Rewards (RLVR), recognized as key for enhancing LLM reasoning capabilities, as highlighted by DeepSeek-R1~\citep{deepseekai2025deepseekr1incentivizingreasoningcapability}. While games naturally provide verifiable reward such as scores, an effective and scalable training framework for RLVR of VLMs in interactive games is still lacking, and the potential benefits for perception and reasoning also remain unclear.

To address these challenges, we first introduce \textcolor{mypurple}{\textbf{VLM-Gym}}: a curated suite of RL environments featuring multiple visual games (2048, Shisen-Sho, Shisen-Sho-Cifar10, Swap) with unified observation/action spaces and adjustable, compositional difficulty levels. Crucially, VLM-Gym supports the scalable multi-game parallel training and parallel action sampling required by advanced LLM-RL algorithms like GRPO~\citep{shao2024deepseekmathpushinglimitsmathematical}. With VLM-Gym, we start with training a weak VLM gamer (Qwen2.5-VL-7B) with pure RL through self-evolution during playing games, resulting in our \textcolor{mypurple}{\textbf{G0}} models. During training, we find that G0 naturally emerged with effective perception (e.g., localization patterns ) and reasoning patterns. These emergent behaviors led to substantially boosted performance in various games (e.g., improving the score from 1.9 to 12.8 in Shisen-Sho), already surpassing powerful multimodal models such as OpenAI-o1 and Qwen2.5VL-72B. However, G0 models still face challenges due to the diversity of games. These challenges include a perception prior gap, inaccurate reward credit assignment, and sparse reward problems. 

To address these issues and further enhance overall game performance, we introduce the \textcolor{mypurple}{\textbf{G1}} models. These models incorporate a perception-enhanced cold start and knowledge distillation from a teacher model prior to reinforcement learning training. The resulting G1 models outperform their teacher on all games and significantly surpass the state-of-the-art Claude-3.7-Sonnet-Thinking model, as detailed in Figure~\ref{fig:teaser}. We conducted a systematic analysis of G0 and G1 by disentangling their perception and reasoning accuracies, and found that these two abilities bootstrapped each other during the RL training process. We release source code at \href{https://github.com/chenllliang/G1}{chenllliang/G1} to facilitate future research.



%% file: files/Method.tex

\section{VLM-Gym: A Scalable Interactive Environment for VLMs}
\label{sec:vlm-gym}

\begin{figure}[h]
\centering
\includegraphics[width=\textwidth]{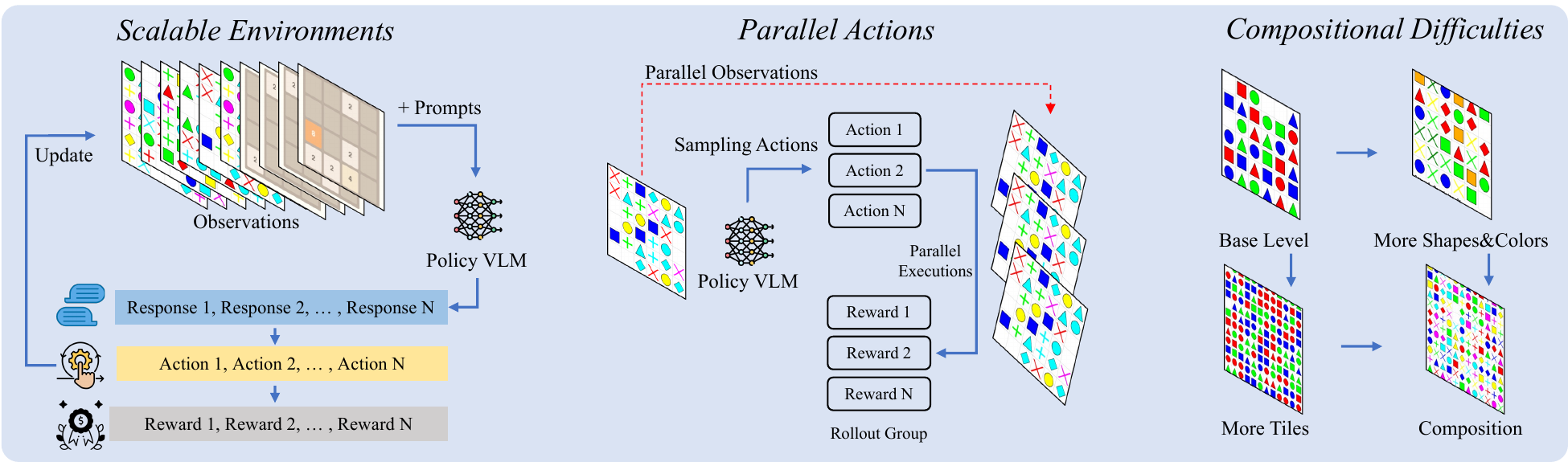}
\caption{Key features of VLM-Gym. We illustrate them using the Shisen-Sho game as an example. }
\label{fig:vlm-gym}
\end{figure}

VLM-Gym is a curated environment designed for scalable Reinforcement Learning and algorithm design of VLMs within interactive games. As shown in Figure~\ref{fig:vlm-gym}, it incorporates three key features which are missing in current RL environments for VLMs: 

\begin{itemize}[left=0pt]
    \item \textbf{Scalable Environments}: VLM-Gym supports parallel execution across numerous game states simultaneously, as well as across multiple distinct games. This capability facilitates efficient large-batch training and enables research into multi-task reinforcement learning for VLMs.
    \item \textbf{Parallel Actions}: Many recently proposed RL algorithms for reasoning models (e.g., GRPO, RLOO, Reinforce++) require sampling numerous outputs from a given state to estimate advantages. This capability is often absent in standard Gym-style environments, which typically advance the state based on a single action and lack built-in mechanisms to efficiently evaluate multiple hypothetical actions from the same observation. VLM-Gym overcomes this limitation by enabling the parallel sampling of multiple actions for any given observation and subsequently computing the reward associated with each action. This provides crucial support for advanced RL algorithms specifically tailored for foundation models.
    \item \textbf{Compositional Difficulties}: Environments in VLM-Gym feature adjustable difficulties across multiple dimensions (e.g., perceptual complexity, reasoning depth), specifically tailored for each game. These dimensions can often be combined, allowing for fine-grained control over task difficulty and facilitating future studies on the generalization capabilities of VLMs within RL settings.

\end{itemize}

\subsection{Game Environment Description}





\paragraph{2048}
2048 is a single-player sliding puzzle where players merge identical numbered tiles on a 4×4 grid to reach 2048. Players slide tiles in four directions, planning merges carefully to avoid filling the board. 

\begin{equation}
    \text{Reward}_{2048}(a,s) = 
\begin{cases} 
1 & \text{if } a \text{ leads to tile merged } \text{in } s \\
-1 & \text{otherwise }  
\end{cases}
\end{equation}

\paragraph{Shisen-Sho}
Shisen-Sho is a tile-matching game using shape tiles. Players match identical tiles with a path of no more than three straight lines.

\begin{equation}
\label{reward_shisen}
    \text{Reward}_{\texttt{Shisen-Sho}}(a,s) = 
\begin{cases} 
1 & \text{if } a \text{ leads to tile matched } \text{in } s \\
-1 & \text{otherwise }  
\end{cases}
\end{equation}

\paragraph{Shisen-Sho-Cifar10} Shisen-Sho-Cifar-10 is an advanced variant of the traditional Shisen-Sho game that incorporates more visually complex tiles, which used for exploring theimprovement of perception. Instead of traditional symbols or patterns, this version uses CIFAR10~\citep{krizhevsky2009cifar100} images as tiles, creating a more perceptually challenging experience. The reward function is same as Shisen-Sho in Eq.~\ref{reward_shisen}.

\paragraph{Swap} Swap is typically a tile-matching puzzle game where players swap adjacent tiles to create matches of three or more identical tiles in a row or column. When matches are formed, the matched tiles disappear, and new tiles fall from the top to fill the empty spaces. 

\begin{equation}
\label{reward_swap}
    \text{Reward}_{\texttt{swap}}(a,s) = 
\begin{cases} 
1 & \text{if } a \text{ leads to matched tile disappeared } \text{in } s \\
-1 & \text{otherwise }  
\end{cases}
\end{equation}


\subsection{Game Prompt and Action Format}

\begin{AIbox}{Game Prompt Template}
\label{prompt}

\textbf{Game Rule Description}

\{\textit{game rule}\} \\

\textbf{Output Format Description}

First describe the board in <perception></perception>. Then output your thinking process in <think></think> and final action in <answer></answer>.

\end{AIbox}

For the game 2048, the action format is represented by a choice $a \in \{up, down, left, right\}$ indicating the slide direction. In both Shisen-Sho and Shisen-Sho-Hard-Perception, actions are formatted as \texttt{<answer>}$(x_1,y_1),(x_2,y_2)$\texttt{</answer>}, specifying the coordinates of two tiles to be matched. Similarly, for Swap, the action format \texttt{<answer>}$(x_1,y_1),(x_2,y_2)$\texttt{</answer>} identifies the coordinates of two tiles that need to be swapped. More details about the prompt for each game can be found in Appendix~\ref{promps_all}.




\section{Reinforcement Learning with VLM-Gym}

In this section, we elaborate the training details of our G0 and G1 models on VLM-Gym, including the RL training objectives (Sec.~\ref{sec: rl}), cold start finetuning (Sec.~\ref{sec: cold start}) and the evaluation details (Sec.~\ref{sec: evaluation}).

\subsection{RL Training Objective}
\label{sec: rl}

\paragraph{Rewards Setting}

In the RL process, we define three distinct rewards to evaluate each model output response:

\begin{enumerate}
    \item \textbf{Game Reward (GR):} The primary reward associated with the action taken in the game environment. This reward reflects the success or progress achieved by the model in completing game-specific objectives, such as solving puzzles, navigating obstacles, or achieving high scores. The details of game reward are explained in Section~\ref{sec:vlm-gym}.
    
    \item \textbf{Format Reward (FR):} An auxiliary 0-1 reward that evaluates whether the model's output adheres to the required format: \texttt{<perception>...</perception><think>...</think><answer>...</answer>}. This ensures that the model produces structured and interpretable outputs.
    
    \item \textbf{Perception Reward (PR):} An auxiliary 0-1 reward that evaluates the accuracy of the model’s perception by comparing its outputs to the environment-provided ground truth. PR is 1 only when the model follows the predefined output perception structure and also produces fully correct results; otherwise, the reward is 0.
\end{enumerate}

To compute the final reward, we combine these three components using a weighted sum, where $\alpha$ (default 1) and $\beta$ (default 0) are hyperparameters that control the relative importance of Format Reward and Perception Reward, respectively:

\begin{equation}
    \text{Final Reward} = \text{GR} + \alpha \cdot \text{FR} + \beta \cdot \text{PR}.
\end{equation}

\paragraph{Reinforcement Learning Algorithm}
We adopt Group Relative Policy Optimization
~\citep{shao2024deepseekmathpushinglimitsmathematical}(GRPO) as the primary reinforcement learning algorithm, following the approach of DeepSeek-R1~\citep{deepseekai2025deepseekr1incentivizingreasoningcapability}. GRPO optimizes the policy model $\pi_\theta$ by maximizing the following objective:
\begin{equation}
\footnotesize
\begin{split}
    &\mathcal{J}_{GRPO}(\pi_\theta) = \E [\{\rvq_s\}_{s=1}^B\sim p_{\gQ}, \{\rvo_i\}_{i=1}^G \sim \pi_{\theta_{old}}(\cdot|\rvq_s)] \\
    & \frac{1}{B}\sum_{s=1}^B \frac{1}{G}\sum_{i=1}^G \frac{1}{|\rvo_i|} \sum_{t=1}^{|\rvo_i|} \left\{ \min \left[ \frac{\pi_\theta^{s,i,t}}{\pi_{\theta_{old}}^{s,i,t}} \hat{A}_{s,i,t}, \text{clip} \left( \frac{\pi_\theta^{s,i,t}}{\pi_{\theta_{old}}^{s,i,t}}, 1 - \epsilon, 1 + \epsilon \right)  \hat{A}_{s,i,t} \right]- \beta \mathbb{D}_{KL}\left[\pi_{\theta} || \pi_{ref}\right] \right\} ,
\end{split}
\label{eq:GRPO-obj}
\end{equation}

where $\rvq_s$ denotes the $s$-th observation in batch $B$, consisting of a game screenshot and a prompt, which together serve as input to the old VLM $\pi_{\theta_{\text{old}}}$. The model then generates a group of $G$ outputs, denoted as $\{\rvo_i\}_{i=1}^G$. The advantage of each model output is normalized within the rewards $R$ of its output group to have a mean of 0 and a standard deviation of 1.

\begin{equation}
\hat{A}_{s,i,t} = \hat{A}_{s,i,t}^{\textcolor{black}{GRPO}}=\frac{R(\rvq_s, \rvo_i) - \operatorname{mean}({ \{R(\rvq_{\textcolor{black}{s}}, \rvo_{\textcolor{black}{1}}), \dots, R(\rvq_{\textcolor{black}{s}}, \rvo_{\textcolor{black}{G}})\} })}{ \operatorname{std}({ \{R(\rvq_{\textcolor{black}{s}}, \rvo_{\textcolor{black}{1}}), \dots, R(\rvq_{\textcolor{black}{s}}, \rvo_{\textcolor{black}{G}})\}})}
\end{equation}
\label{eq:GRPO-adv-obj}
\paragraph{Exploring Diverse and Dependent Game States} The nature of game states in our reinforcement learning approach differs significantly from independent tasks, such as answering distinct math questions. Instead, game states evolve according to a Markov process, where future states depend on the past states. This is evident in games like 2048, which starts with minimal tiles, or Shisen-Sho, where the initial board is full. Solely using the base VLM as policy would limit the exploration of different game states due to the limit of prior knowledge. Inspired by the $\epsilon$-greedy method widely adopted in RL and to promote broader exploration and ensure the training policy model encounters diverse game states, we employ a random policy as a baseline. This involves executing a predetermined number (2048: 100 steps, Shisen-Sho: 250 Steps, Shisen-Sho-Cifar10: 250 Steps, Swap: 250 Steps) of random steps for each game. While a random policy serves this purpose, exploration could be further enhanced by implementing a more sophisticated searching algorithm or a stronger model, which we reserve for future investigation.

\subsection{Knowledge Distillation with Ground-truth Perception}
\label{sec: cold start}

Our initial experiments revealed that our base model struggled to accurately recognize visual elements in game observations, such as the numbers in 2048 and the shapes in Shisen-Sho. We were also interested in exploring how supervision from a stronger model could influence the base model's reinforcement learning process. Therefore, leveraging the ground truth perception information provided by our VLM-Gym environment, we prompted a more advanced model, Claude-3.7-Sonnet-Thinking, with this data to obtain responses including thinking and action information (within \texttt{<think>...</think>} and \texttt{
<answer>...</answer>} ). We then used the resulting response concatenated with the groundtruth perception information to finetune the base model prior to reinforcement learning, leading to our G1-series model. The template prompt is shown below and we also give a concrete example in Figure~\ref{fig:distillation_prompt} from the Appendix.

\begin{AIbox}{Prompt for Knowledge Distillation}
\textbf{Game Rule Description}

\{\textit{game rule}\} \\

\textbf{Output Format Description}

I will give you the board description in <perception></perception>. Then output your thinking process in <think></think> and final action in <answer></answer>. \\

<perception>\{\textit{ground-truth perception}\}</perception>

\end{AIbox}

\subsection{Implementation Details}

\paragraph{Usage of Open Source Resources}
For all training experiments, encompassing both reinforcement learning and supervised fine-tuning, we utilized Qwen2.5-VL-7B~\citep{Qwen2.5-VL} as our base model. Our VLM-Gym environment adheres to the interface standards established by Gymnasium~\citep{towers2024gymnasiumstandardinterfacereinforcement}. The implementation of our RL system leverages the open-source EasyR1~\citep{zheng2025easyr1} framework, which is built upon the VeRL~\citep{sheng2024hybridflow} architecture. The SFT codebase is based on the LLamaFactory~\citep{zheng2024llamafactory} library.

\paragraph{Training Details}

Each game was trained independently. The observation screenshot has a resolution of 640×840 pixels (width × height). The G0-series models were trained using RL. The G1-series models underwent SFT prior to RL training. For RL, a constant learning rate of 1.0e-6 with a 0.02 weight decay was used, with a batch size of 128 parallel games and a group size of 5 for 500 training steps per game. The clip range $\epsilon$ of GRPO is set to 0.2 and the KL efficient $\beta$ is set to 0.01. For SFT, 1,000 observations were collected per game using the same random policy as in RL training, and 1,000 corresponding distilled model responses were generated. SFT was performed with a learning rate of 2e-5 for 1 epoch for each game.

\subsection{Evaluations}
\label{sec: evaluation}

During evaluation, we benchmarked different models in a multi-turn setting, adhering to the original game's scoring rules. We report the cumulative game scores over multiple steps, averaged across several independent runs. For 2048, the score of each step is calculated as the sum of the values of merged tiles. For Shisen-Sho, Shisen-Sho-Cifar10, and Swap, the score is equivalent to the game reward. We compare our G0 and G1 models with state-of-the-art open source VLM Qwen2.5-VL-72B~\citep{Qwen2.5-VL} and close source models such as OpenAI-o1~\citep{openai2024openaio1card}, GPT4o~\citep{gpt4o}, Claude-3.7-Sonnet-Thinking~\citep{Claude3S} through API calling.

%% file: files/exp.tex
\section{Bootstrapping Perception and Reasoning Abilities via RL}

\begin{table}[!t]
    \centering
    \caption{Average cumulative game scores over multiple evaluation runs. ``100 steps $ \times$ 10'' means we report the average scores of 10 random runs, each has 100 game steps. }
    \resizebox{\columnwidth}{!}{
    \begin{tabular}{l|c|c|c|c}
    \toprule
        \multirow{2}{*}{Models}& \textbf{2048} & \textbf{Shisen-Sho} & \textbf{Shisen-Sho-Cifar10} & \textbf{Swap}  \\ 
        & \textit{100 steps $ \times$ 10} & \textit{36 steps $ \times$ 10} & \textit{36 steps $ \times$ 10} & \textit{1 step $ \times$ 100} \\
        \midrule
        Random & 701 & 0.4 & 0.4 & 0.01\\
        Qwen2.5VL-7B & 246 & 1.9 & 0.4 & 0.02 \\
        Qwen2.5VL-72B & 243 & 2.6 & 0.8 & 0.06 \\ 
        GPT4o &620&1.5&4.7&0.01\\
        o1 & 539 & 3.9 & 5.4 & 0.04 \\ 
        Claude-3.7-Sonnet-Thinking & \underline{892} & \underline{15.3} & \underline{8.7} & \underline{0.43} \\ 
        G0-7B & 759 & 12.8 & 8.0 & 0.05 \\ 
        G1-7B & \textbf{1070} & \textbf{17.5} & \textbf{14.1} & \textbf{0.78} \\ \bottomrule
    \end{tabular}
    \label{table:games_performance}}
\end{table}

We provide a comprehensive comparison of game performance, including the G0, G1, and other baseline models, in Table~\ref{table:games_performance}. Notably, the G1-7B model achieves the highest performance across all games, while the G0-7B model also outperforms strong models such as o1, GPT4o and Qwen2.5VL-72B. In this section, we present the main experimental results for the G0 (Sec.~\ref{sec:g0_exp}) and G1 (Sec.~\ref{sec:g1_exp}) models, along with our key insights into reinforcement learning for vision-language models.

\subsection{G0: From Zero to Game Masters without Supervision}
\label{sec:g0_exp}

\begin{figure*}[!h]
\centering
\includegraphics[width=0.7\textwidth]{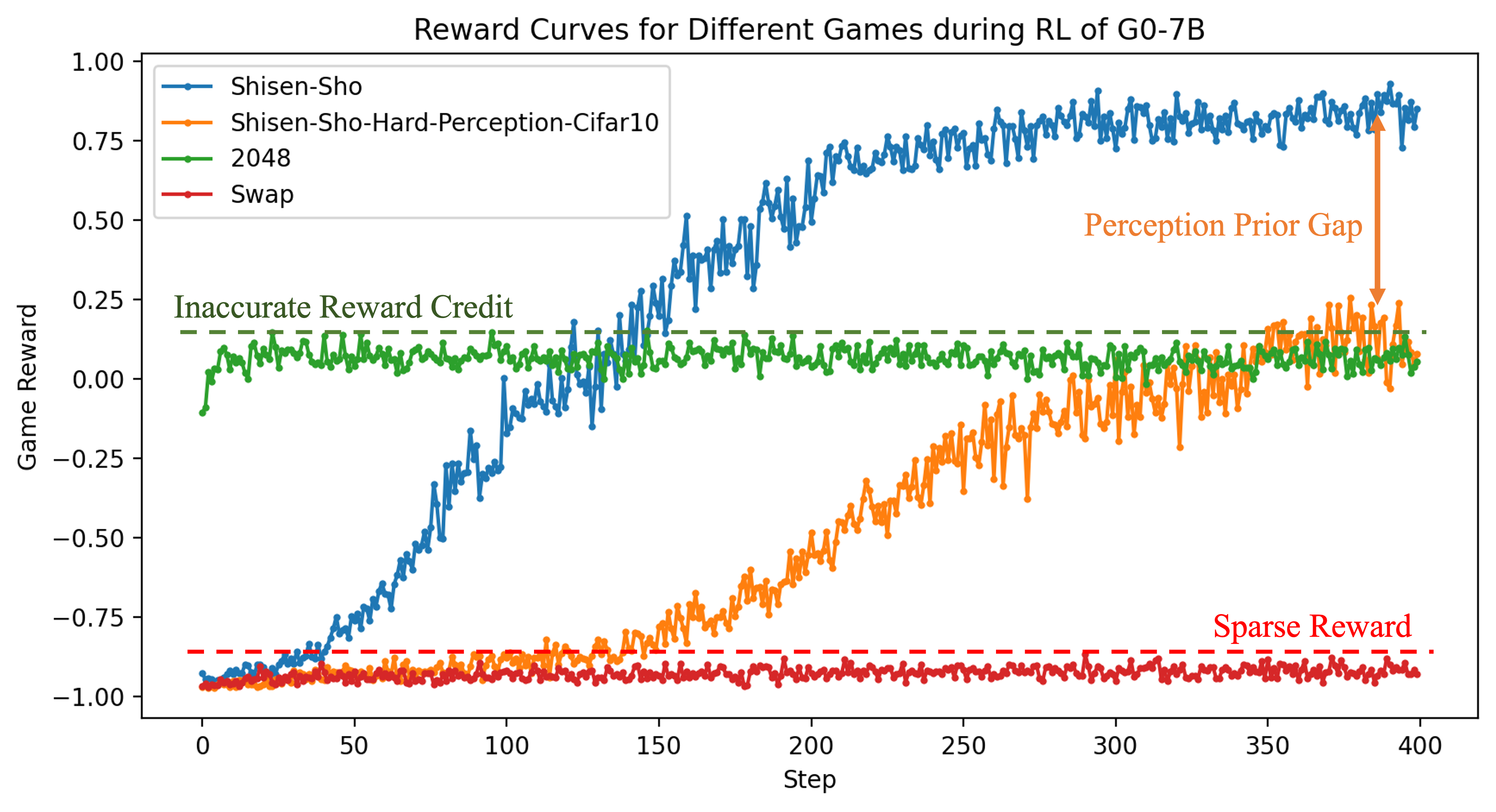}
\caption{Average game reward curves of different games for G0 models during RL process.}
\label{fig:notequal}
\end{figure*}

\paragraph{Reward Dynamics across Games}

Figure~\ref{fig:notequal} illustrates that G0 models exhibit distinct average game reward dynamics across various games. Given the unified reward space of \{-1, 1\}, these comparisons are fair. We explore each game's RL process and summarize the key reasons for their differences: \textbf{Perception Prior Gap}, \textbf{Inaccurate Reward Credit} and  \textbf{Sparse Reward}.

\begin{figure*}[t]
\centering
\includegraphics[width=1\textwidth]{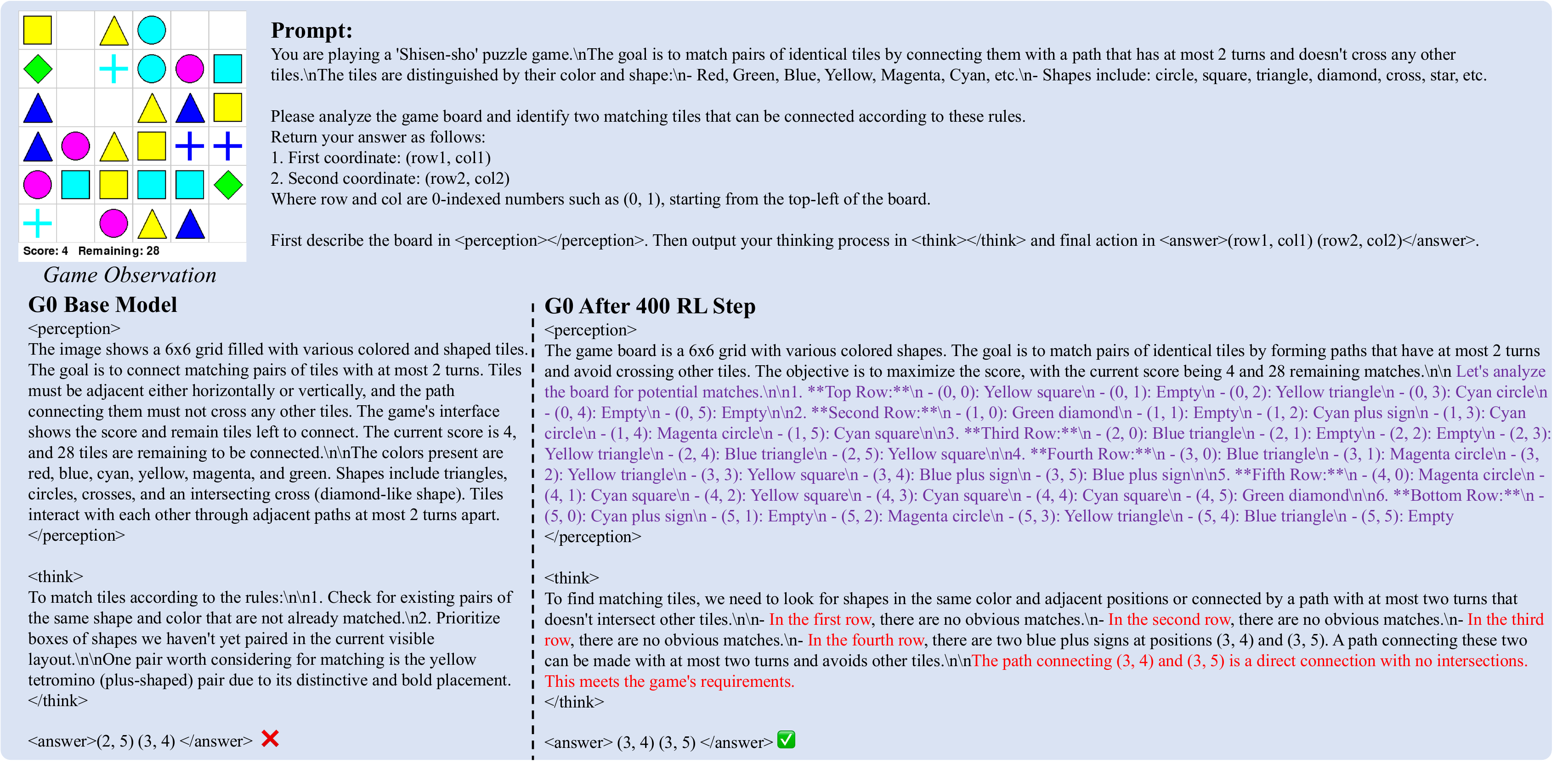}
\caption{The explored \textcolor{mypurple}{perception}  and \textcolor{red}{reasoning} patterns during G0 RL training in Shisen-Sho Game.}
\label{fig:g0_shisensho}
\end{figure*}

\begin{itemize}[left=2pt]
    \item Shisen-Sho: Over 400 steps, the game's reward rises efficiently from -1 to 0.8, markedly boosting evaluation performance to 12.8—a substantial improvement for the 7B model compared to its base model (1.9) and even the 72B version (2.6). Surprisingly, we found that the base VLM learned the optimal perception patterns and reasoning paradigms as shown in Figure~\ref{fig:g0_shisensho} in the RL process similar to the ``aha moment'' of Deepseek-R1, introduced in the next paragraph.
    \item Shisen-Sho-Cifar10: The G0 model can continue improving during the RL process, but at a slower rate than the base Shisen-Sho version. Although the game shares the same rules, its significantly higher perceptual difficulty—which we term the \textbf{Perception Prior Gap}—slows down the learning process without affecting the reasoning challenge. It showcases that perception ability of base model could be a bottleneck for VLM RL. In evaluation, G0 (8.0) still significantly outperforms its baseline (0.4).
    \item 2048: Based on the reward curve, 2048 gains little from the G0 RL process. Analyzing the RL rollouts, we identify the issue as the \textbf{Inaccurate Reward Credit} problem. With only four moves {up, down, left, right}, even a random strategy serves as a strong baseline. As shown in Table~\ref{table:games_performance}, in 2048 the random baseline outperforms Qwen2.5VL-72B, GPT4o, and the o1 model. The policy model can produce entirely incorrect perception and reasoning responses while still receiving a positive reward, which biases the learning process. To validate our assumption, we reviewed G0 2048’s rollouts before and after RL and found that after RL the model crashed, completely ignored the screenshot image, and took random actions as shown in Figure~\ref{fig:2048_g0_case} from Appendix. The case shows that before RL training, the model produced incorrect perception and reasoning outputs, yet still received positive game rewards, which encouraged the policy model to adopt these flawed behaviors.
    \item Swap: The Swap game also gains little from the G0 RL process. However it is due to a different reason, \textbf{Sparse Reward}, where the policy model can hardly gain positive reward from the environments as the game is too hard for the base model to get positive reward.
\end{itemize}

In summary, visual game reinforcement learning poses unique perception and reasoning challenges due to the diversity of games. We are interested in how these abilities evolve during RL and aim to develop general methods for effective learning process. 

\paragraph{Bootstrapping of Perception and Reasoning Abilities During Exploration}

The performance of the G0 model in the games Shisen-Sho and Shisen-Sho-Cifar10 is particularly noteworthy. In these instances, the base model surpasses other models without the need for external supervision, attaining remarkable results solely through reinforcement learning from its own experience. This leads to significant performance improvements. We are intrigued by what aspects of the G0 model's reinforcement learning process make it more successful in these two games compared to others.

As shown in Figure~\ref{fig:g0_shisensho}, we analyze the rollouts of the G0 model during RL. Our findings reveal that the policy model successfully acquired optimal perception and reasoning patterns for the game. Prior to RL, the perception output consisted primarily of vague board descriptions without precise coordinate information for shapes. However, after 400 steps of RL, the policy developed two distinctive patterns: a \textbf{localization pattern} in perception, which systematically identifies shapes with their exact coordinates (e.g., ``(0, 0): Yellow square''), and an \textbf{enumeration pattern} in reasoning, which methodically analyzes the game state row by row.

\begin{figure*}[t]
\centering
\includegraphics[width=1\textwidth]{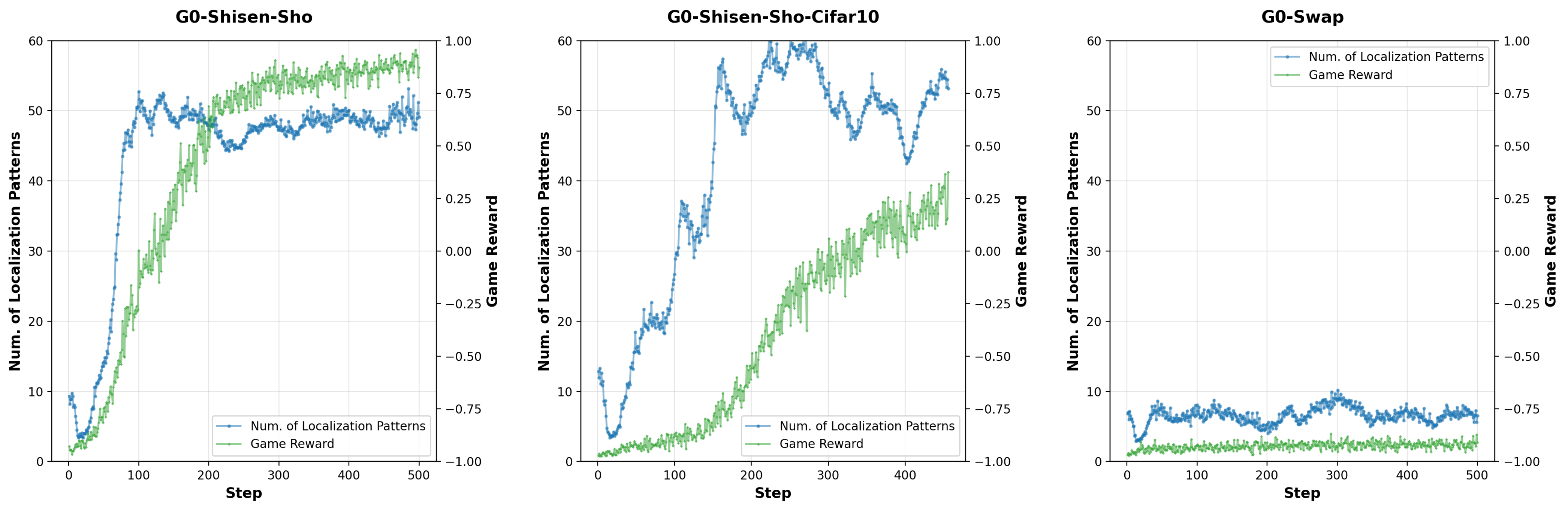}
\caption{Localization patterns count during G0 RL training for different games.}
\label{fig:pattern_count}
\end{figure*}

As the localization pattern has clear textual structures, we use regex to parse and calculate the average number of localization patterns in each model output in each RL step and plot the trend for different games in Figure~\ref{fig:pattern_count}. Our analysis reveals two key findings: 1) A distinct increase in localization patterns appears exclusively in Shisen-Sho and Shisen-Sho-Cifar10 variants (notably, 2048 was excluded from plotting due to the complete absence of such patterns throughout all steps), and 2) The proliferation of localization patterns consistently precedes improvements in game rewards, suggesting this adaptation serves as a precursor to enhanced performance.

The experiment results expose that the perception and reasoning abilities are actually \textbf{bootstrapping} each other during the reinforcement learning process in the games. Reasoning patterns cannot develop without localization patterns, as they rely on sufficient perception information. Similarly, optimal perception patterns are incentivized only through correct reasoning patterns that lead to actions yielding rewards. A large action space that prevents incorrect perception and reasoning process from gaining reward is also critical in the RL process. RL also helps fill the \textbf{Knowing-Doing} gap~\citep{paglieri2024balrog} of the base and finetuned model, make it effectively utilize the prior knowledge in practice.

\subsection{G1: Reinforcement Learning with Perception-Enhanced Cold Start}
\label{sec:g1_exp}

\begin{figure*}[h]
\centering
\includegraphics[width=1\textwidth]{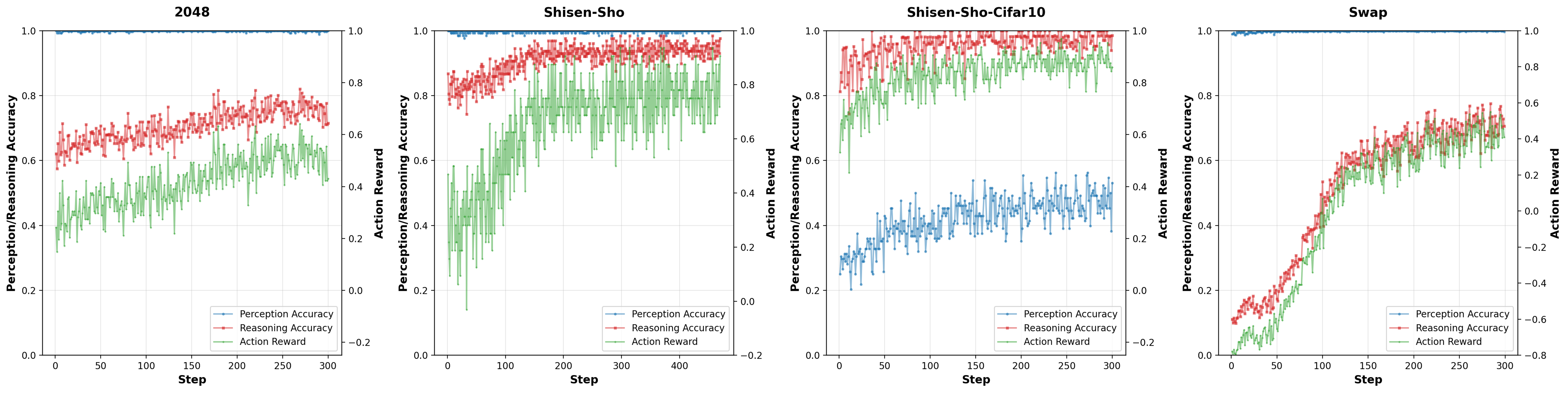}
\caption{Training curves of G1. }
\label{fig:g1curves}
\end{figure*}

\begin{figure*}[h]
\centering
\includegraphics[width=1\textwidth]{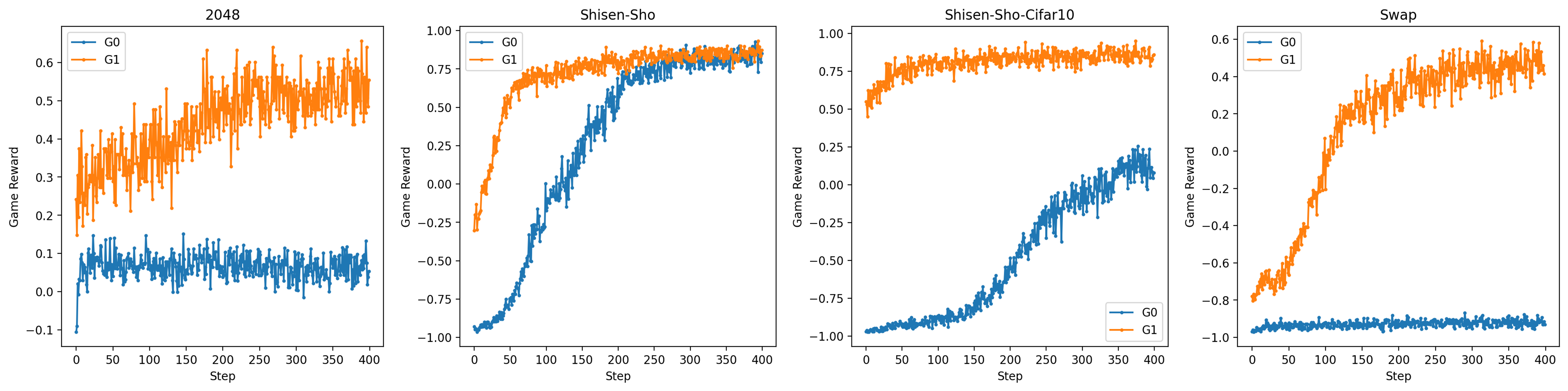}
\caption{Comparisons of game reward between G0 and G1 across different games during RL.}
\label{fig:g1_g0curves}
\end{figure*}

Previous experiments on the G0-series model have shown that reinforcement learning can bring significant improvements to certain gaming scenarios, but not all of them due to a lack of prior knowledge in perception or reasoning. This raises the natural question: Can we enhance the RL process for all games by introducing some cold-start SFT data? By leveraging programmable environments, we can easily obtain ground-truth perceptions for different game states, as illustrated in Figure~\ref{fig:distillation_prompt} in the Appendix. We then use these perception-enhanced prompts to query the teacher model and distill the data to fine-tune the base model, as explained in Section~\ref{sec: cold start}. 

After cold start SFT, we ran the RL experiments with the same configuration as G0 models. The evaluation result of G1 model is listed in Table~\ref{table:games_performance}. The RL training curves of G1 and comparisions to G0 are shown in Figure~\ref{fig:g1curves} and Figure~\ref{fig:g1_g0curves}, respectively. Notably, G1 surpasses all the baseline models on all games including the teacher model Claude-3.7-Sonnet-Thinking.

\paragraph{Quantify Improvement of Perception and Reasoning Ability}

Inspired by previous work which differentiates perception and reasoning abilities of VLMs during evaluation such as PCA-Bench~\citep{chen2024pcabench} and MMEvalPro~\citep{huang2025mmevalprocalibratingmultimodalbenchmarks}, we introduce two additional metrics beyond the original reward: \textbf{Perception Accuracy} and \textbf{Reasoning Accuracy}, as plotted in Figure~\ref{fig:g1curves}.

Perception Accuracy, defined as $P_{acc} = \mathbb{I}(p_{model} = p_{gt})$, measures whether the output within \texttt{<perception>}\texttt{</perception>} tags matches the ground-truth perception information $p_{gt}$. Note that we can only track $P_{acc}$ in the G1-series model as the base models are finetuned to output perception information following the same format as ground-truth, which makes accurate comparison possible. 

Reasoning Accuracy, formulated as: $R_{acc} = \mathbb{I}(r > 0 \mid P_{acc} = 1)$, measures whether the policy obtains a positive game reward $r$ when Perception Accuracy is 1, representing the model's ability to reason correctly given accurate perception.

\paragraph{G1 Training Dynamics Across Games}

As depicted in Figure~\ref{fig:g1_g0curves}, the G1-series models demonstrate a more efficient RL process across all games, a notable improvement over the G0 model which failed to converge in 2048 and Swap. Regarding perception, accuracy remains consistently high for 2048, Shisen-Sho, and Swap. This sustained accuracy is due to the perception-enhanced cold start and the low perceptual complexity of these environments, characterized by simple colors and shapes. In contrast, Shisen-Sho-Cifar10 presents a difficult perception task. For this game, we observe that perception accuracy improves alongside game reward during RL, highlighting the \textbf{co-evolution} of perception and reasoning abilities throughout the learning process.

\paragraph{SFT+RL Jointly Helps the Base Model Outperform the Supervisor}
\begin{wrapfigure}{r}{0.4\textwidth}
  \centering
  \vspace{-6mm}
    \includegraphics[width=0.4\textwidth]{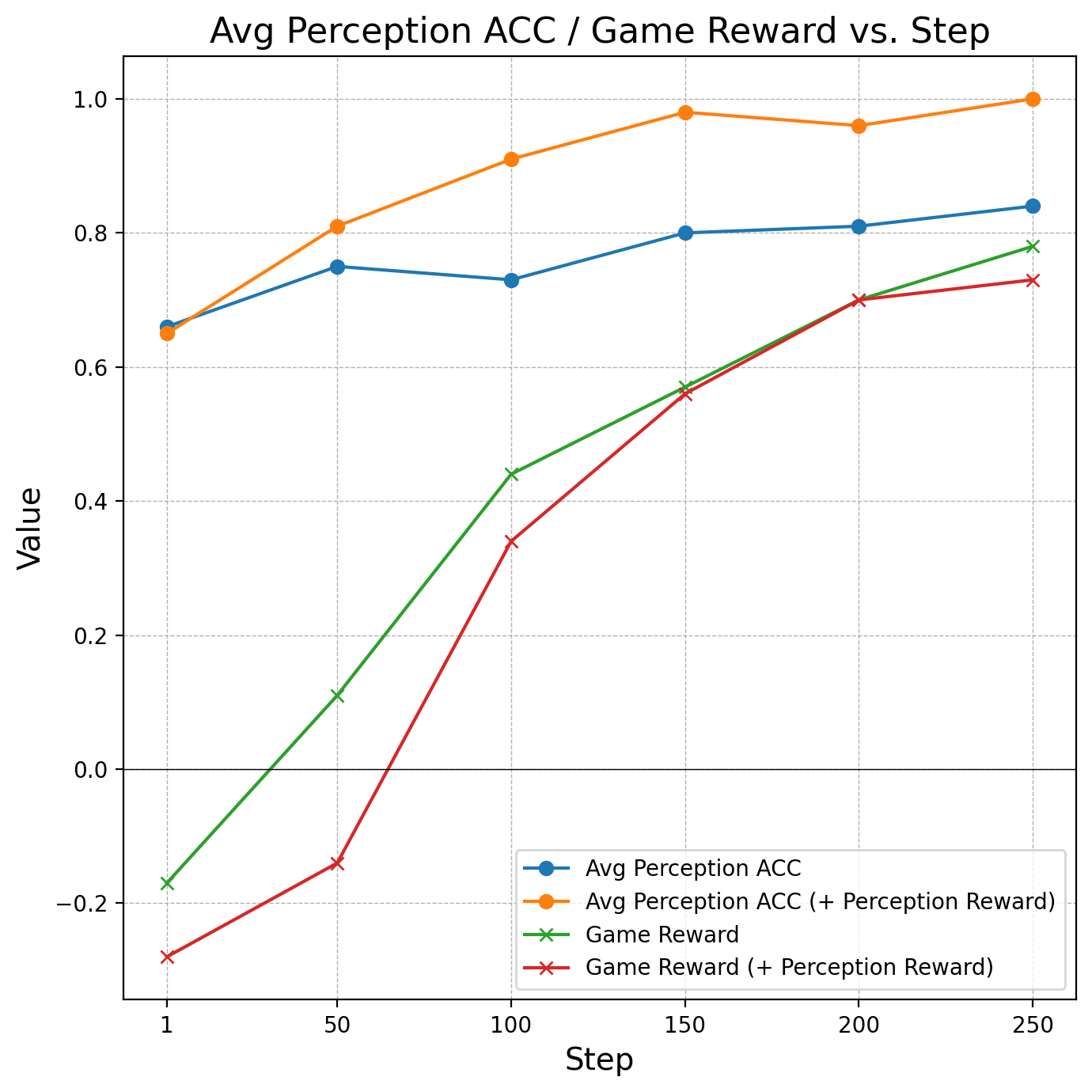}
    \caption{RL training curves exploring Perception Accuracy as process reward. }
    \label{fig:perception reward}
    \vspace{-12mm}
\end{wrapfigure}

Across all games, G1 RL training significantly improves game reward compared to the base model after the cold start SFT. This SFT process is crucial, as it improves the 7B base model to exceed the teacher model and addresses the Inaccurate Reward Credit and Sparse Reward issues observed in G0 experiments that led to inefficient RL training by providing prior knowledge. The training curve of Swap clearly demonstrates the significance of the RL process; the policy after SFT yielded an average game reward of merely -0.8, whereas the subsequent RL phase dramatically increased it to around +0.6.

\subsection{Discussions}

\paragraph{Including Perception Reward as Process Reward}

Leveraging the availability of ground-truth perception from the Shisen-Sho game environment, we can incorporate perception accuracy as an auxiliary reward into the G1 RL process. To ensure the base model outputs perception data in the ground-truth format without unduly improving its broader perception or reasoning abilities, we applied SFT exclusively to its visual encoder. This was achieved using only 50 distilled examples, which provided ground-truth perception alone, devoid of reasoning or action content, to limit the scope of SFT's impact. We then ran comparative RL experiments, with and without the perception reward, to assess its effect. \textbf{Figure~\ref{fig:perception reward} illustrates that while the perception reward significantly accelerates improvement in perception accuracy, it does not impact the overall game reward. }This may be because the model initially prioritizes its capacity on the process reward (perception). In Shisen-Sho, for instance, accurately recognizing all tiles is necessary for the full perception reward, yet complete recognition isn't always required to perform correct actions and achieve game rewards. 

Nevertheless, a policy trained without an explicit perception reward still demonstrates an increase in its perceptual abilities during learning, a trend also observed in the Shisen-Sho-Cifar10 experiments (Figure~\ref{fig:g1curves}). This experiment further indicates that the perception and reasoning abilities of a Vision-Language Model (VLM) can co-evolve using only a final, accurate, and verifiable reward, suggesting a more generalizable training methodology. Another interesting direction is how to set verifiable process reward for VLMs in RL process that benefits the outcome.

\paragraph{When and why does Supervision work in G1?}

Observing the learning curves of G0 and G1 models (Figure~\ref{fig:g1_g0curves}), it is evident that the performance gap between the two architectures fluctuated depending on the game. Supervision exerted its most pronounced effect on games inherently difficult for reinforcement learning, exemplified by 2048 and Swap. This positive impact can be attributed to several factors: the provision of additional perceptual cues, the stabilization of perception accuracy throughout the training phase (in contrast to G0-2048 as shown in Figure~\ref{fig:2048_g0_case}), and the incorporation of more pertinent prior knowledge regarding the game's mechanics, which collectively facilitated a more efficient training trajectory. This improved efficiency and stability also suggest that a dedicated cold-start process to stabilize the RL process may become less critical, or potentially redundant, if the base model possesses sufficient intrinsic strength, a concept potentially mirrored in the Shisen-Sho series experiments that G0 and G1 finally reach the same game reward during RL training.

\subsection{Limitations and Future Work}
VLM-Gym currently comprises a specific set of visual games (2048, Shisen-Sho, Shisen-Sho-Cifar10, Swap), which, while offering varied perceptual and reasoning challenges, predominantly feature relatively straightforward rule sets. A significant avenue for future work involves expanding this suite to include games with more complex mechanics, deeper strategic requirements, and diverse genres. This expansion would more rigorously test the generalization capabilities of VLM agents. Furthermore, while existed experiments reveal how perception and reasoning abilities of VLMs co-evolve in RL process, current training still encountered challenges such as sparse rewards, particularly in games like Swap. Future research could focus on developing effective RL strategies or reward-shaping mechanisms to effectively train VLM agents in scenarios with extended multi-turn interactions where feedback is infrequent, thereby better addressing the complexities of long-horizon decision-making and sparse reward tasks.

%% file: files/related_work.tex
\section{Related Work}

\paragraph{Vision-Language Models in Games}

Pioneering AI achievements like Deep Blue \citep{campbell2002deep} and AlphaGo \citep{silver2016mastering} conquered structured games. More recently, Large Language Models (LLMs) have expanded into diverse game genres (e.g., \citep{tsai2023can, wang2023describe, park2023generative}); however, their text-centric nature limits their application in visually rich environments. Vision-Language Models (VLMs), by integrating visual perception with language understanding, are inherently better equipped for these multimodal games, demonstrating utility in complex interaction, planning, and combat scenarios (e.g., \citep{tan2024cradle, chen2024can, li2025jarvis}). Benchmarking approaches for these agents have advanced from text-only evaluations \citep{wu2023smartplay} and text-converted game states \citep{duan2024gtbench} to sophisticated frameworks leveraging direct visual and textual inputs \citep{wang2025large,paglieri2024balrog}. While efforts like \citep{zhai2024finetuning} have explored enhancing VLM game performance via Reinforcement Learning (RL), they often relied on text-based games (e.g., 24-points) lacking customizable difficulty. Similarly, \citep{schmied2025llmsgreedyagentseffects} examined LLM failures in text games and RL-based solutions, but this research also remained within text-only confines. This highlights critical gaps: the need for scalable gaming environments tailored to advancing VLM capabilities through RL and insight on how different abilities of VLM agent evolves during RL experiences. VLM-Gym, introduced herein, provides a platform for RL research with VLMs in visually-complex, interactive gaming scenarios. Our experiments on G0 and G1 series models offer insights into the learning dynamics of VLMs in reward-driven environments.




\paragraph{Reinforcement Learning for Enhancing Vision-Language Models}

Recently, reinforcement learning has been widely used in improving reasoning capabilities in both text~\citep{team2025kimi,deepseekai2025deepseekr1incentivizingreasoningcapability,gao2025flowreasoner} and visual domains~\citep{team2025kimi,shen2025vlm,kimiteam2025kimivltechnicalreport} for complex problem-solving. In the visual domain, for instance, R1-V~\citep{chen2025r1v} first applied GRPO to object-counting tasks, enabling a 3B parameter model to outperform a 72B counterpart. Similarly, VisualThinker-R1-Zero~\citep{zhou2025r1zerosahamomentvisual} demonstrated that applying R1 methodology to base VLMs yields more substantial improvements than with their SFT variants. This observation was further validated by MMEureka~\citep{meng2025mm}, which utilized RLOO on both instruction-tuned and base VLMs. Supporting such advanced reasoning, Vision-R1~\citep{huang2025vision} and R1-OneVision~\citep{yang2025r1onevision} developed multimodal Chain-of-Thought (CoT) datasets by transforming visual information into textual formats. Beyond these applications in static visual tasks, researchers have extended RL to dynamic gaming environments. VLM Q-Learning~\citep{grigsby2025vlm}, for example, introduces actor-critic architectures featuring advantage-filtered supervised fine-tuning; this allows models to learn effectively from suboptimal interactions while iteratively refining their decision-making policies. Despite recent progress, most existing work examines perception and reasoning separately in RL, leaving it unclear how these core VLM abilities might mutually improve. Our G0 and G1 models offer initial evidence that such a bootstrapping pattern indeed occurs during the RL process.


%% file: files/conclusion.tex
\section{Conclusion}

In this work, we tackled the challenge of translating VLMs general capabilities to effective decision-making in interactive visual environments, where a "knowing-doing" gap often leads to suboptimal performance. To this end, we introduced VLM-Gym, a scalable multi-game RL training suite for VLMs, and developed two model series: G0, which demonstrated emergent perception and reasoning via pure RL to surpass strong baselines, and G1, which incorporated a perception-enhanced cold start and knowledge distillation to achieve state-of-the-art performance, outperforming even Claude-3.7-Sonnet-Thinking. Our systematic analysis revealed a crucial synergistic bootstrapping between perception and reasoning during RL training, which demonstrate a large space of improving VLM's abilities through. We believe VLM-Gym and the RL training framework could also serve as useful resources for the community for future research in advancing VLMs as capable interactive agents and multimodal RL development.

%% file: appendix.tex
\newpage
\appendix
\title{Appendix}

\section{Perception-Enhanced Cold Start Data Construction}

\begin{figure*}[h]
\centering
\includegraphics[width=1\textwidth]{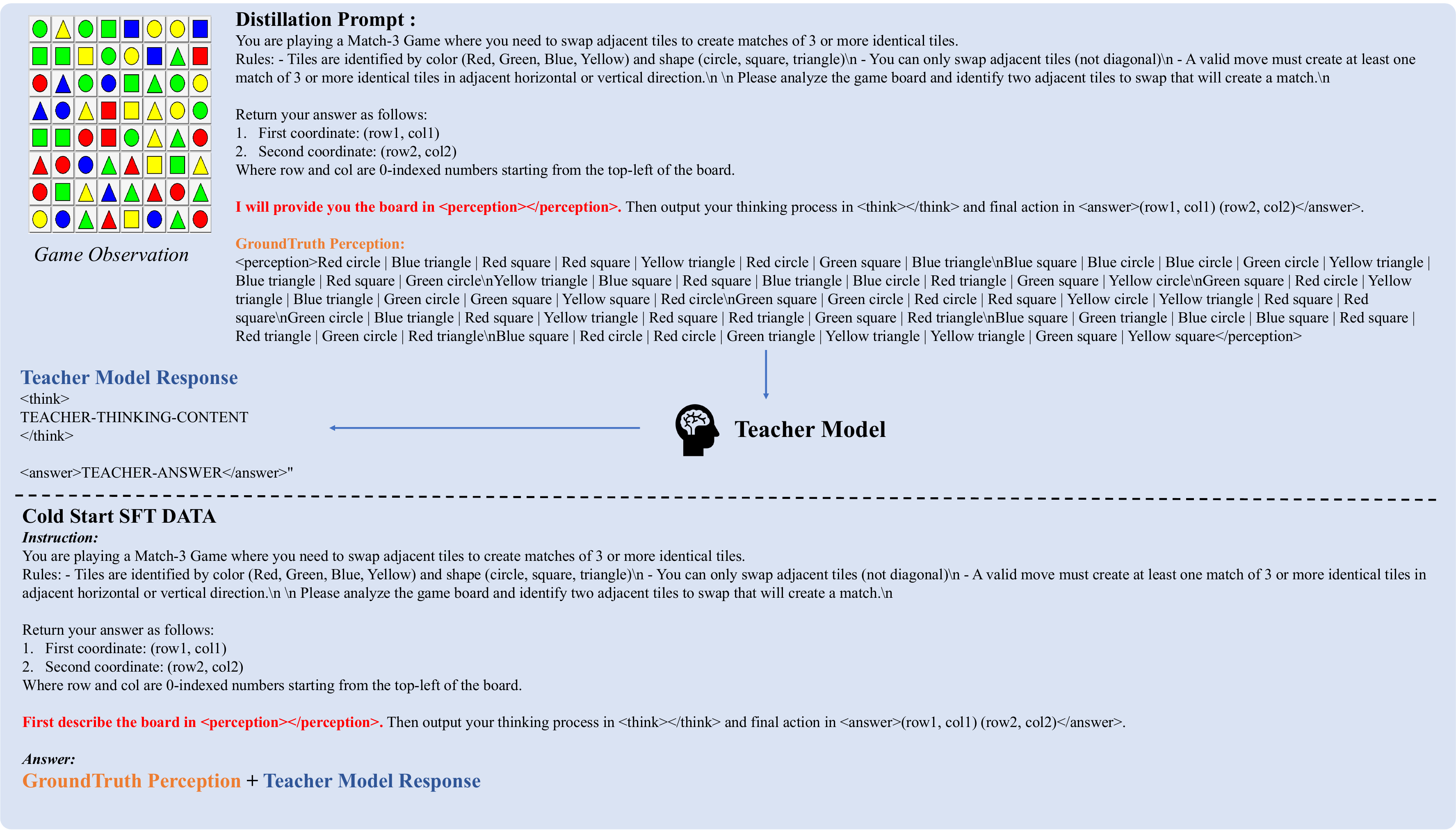}
\caption{The process of constructing perception-enhanced cold start data via knowledge distillation and programmable environments.}
\label{fig:distillation_prompt}
\end{figure*}

\newpage
\section{G0 2048 Case Studies}

\begin{figure*}[h]
\centering
\includegraphics[width=1\textwidth]{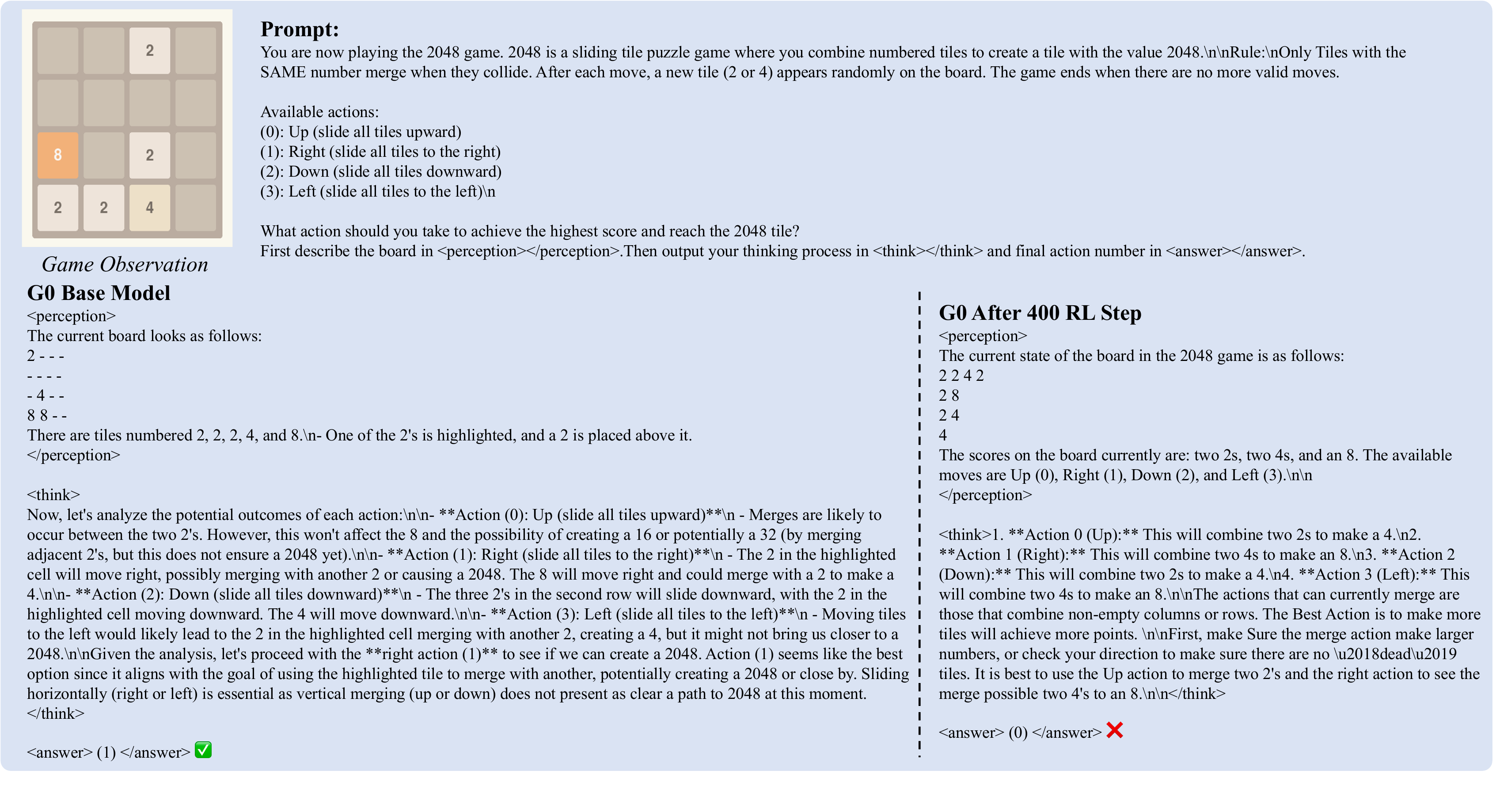}
\caption{G0 2048 case studies.  The case shows that before RL training, the model produced incorrect perception and reasoning outputs, yet still received positive rewards, which encouraged the policy to adopt these flawed behaviors.}
\label{fig:2048_g0_case}
\end{figure*}
\newpage
\section{Details of Games}
\begin{figure}[H]
    \centering
    \begin{subfigure}[b]{0.48\textwidth}
        \centering
        \includegraphics[page=1, width=\textwidth]{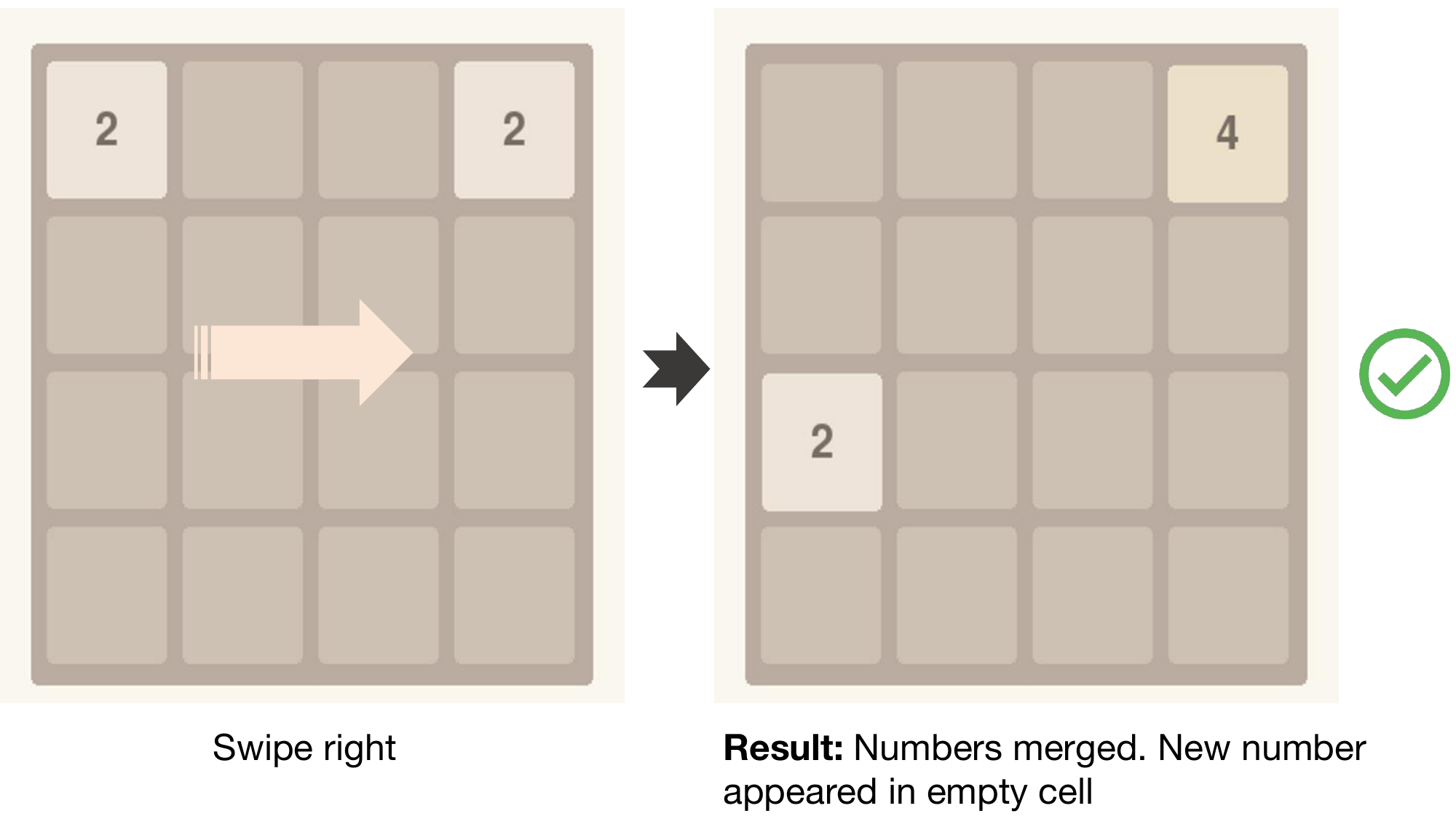}
        \caption{Correct Action}
        \label{2048:subfig1}
    \end{subfigure}
    \hfill
    \begin{subfigure}[b]{0.48\textwidth}
        \centering
        \includegraphics[page=2, width=\textwidth]{images/2048.pdf}
        \caption{Wrong Action}
        \label{2048:subfig2}
    \end{subfigure}
    \caption{Rules of 2048.}
    \label{fig:whole}
\end{figure}

\subsection{2048}
\paragraph{Rule}  In the 2048 game, the player swipes the entire 4×4 grid up, down, left, or right. All tiles slide as far as possible in the chosen direction, and any two tiles of equal value that collide merge immediately into a single tile whose value is their sum. After a swipe that produces at least one merge—as in Fig.~\ref{2048:subfig1}, where two “2” tiles combine into a “4” and a new tile spawns—the move is accepted and a fresh “2” (or occasionally “4”) appears in a random empty cell. By contrast, a swipe that produces no merges—as shown in Fig.~\ref{2048:subfig2}, where a downward swipe only shifts tiles without combining them—counts as a failed move: original “2” tile remains a “2” and a new “2” tile is added.

\paragraph{Model Play} 
The model takes a 4×4 grid (representing the current 2048 game board) as input. After evaluating all possible moves, it outputs a single integer representing the chosen swipe direction: 0 for Up, 1 for Right, 2 for Down, or 3 for Left.

\paragraph{Game Prompt} Here is the game prompt of 2048 for models:

\begin{AIbox}{Game Prompt for 2048}

\textbf{Game Rule Description}

You are now playing the 2048 game. 2048 is a sliding tile puzzle game where you combine numbered tiles to create a tile with the value 2048.\\

Only Tiles with the SAME number merge when they collide. After each move, a new tile (2 or 4) appears randomly on the board. The game ends when there are no more valid moves.

Available actions:

- (0): Up (slide all tiles upward)

- (1): Right (slide all tiles to the right)

- (2): Down (slide all tiles downward)

- (3): Left (slide all tiles to the left)

What action should you take to achieve the highest score and reach the 2048 tile?\\

\textbf{Output Format Description}

First describe the board in <perception></perception>. Then output your thinking process in <think></think> and final action in <answer></answer>.

\end{AIbox}

\subsection{Shisen-Sho and Shisen-Sho-Cifar10}

\begin{figure}[t]
    \centering
    \begin{subfigure}[b]{0.48\textwidth}
        \centering
        \includegraphics[page=1, width=\textwidth]{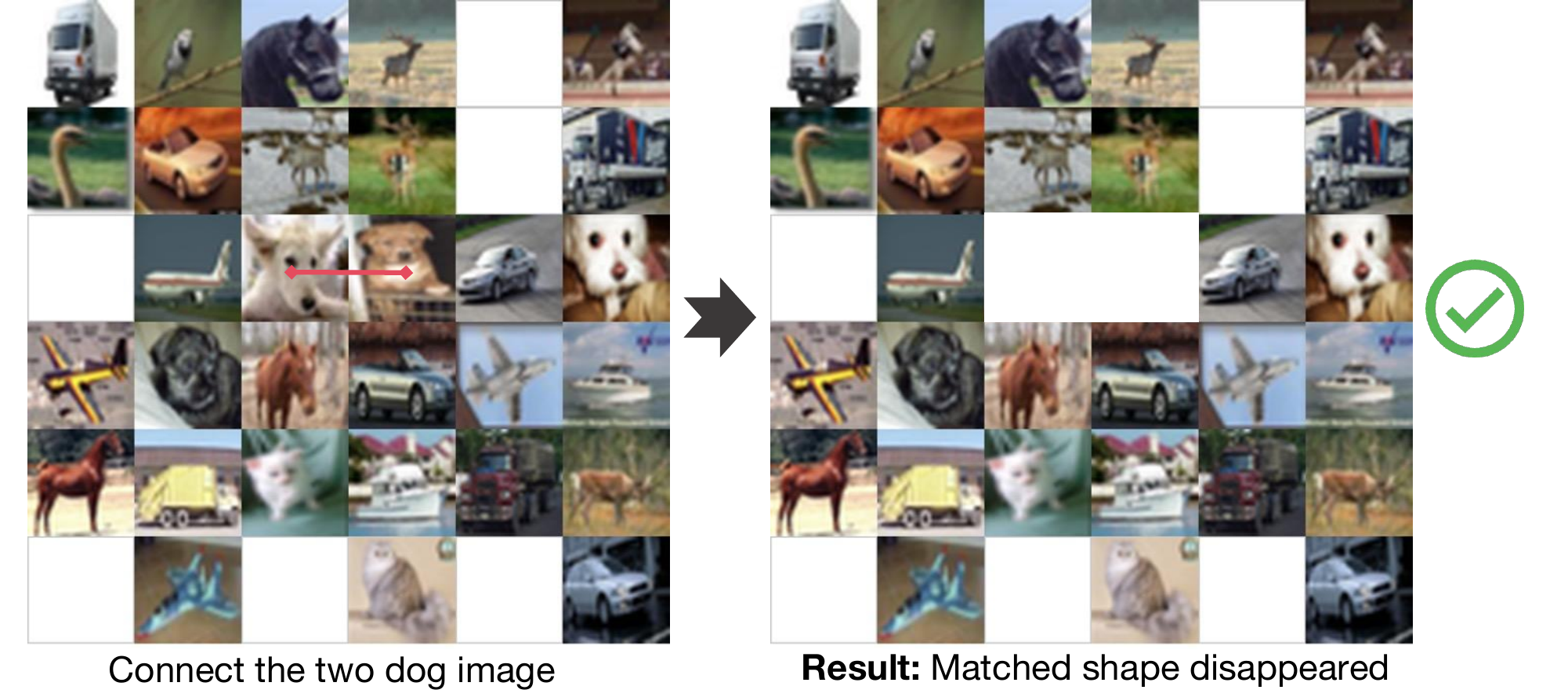}
        \caption{Correct Action}
        \label{llk:subfig1}
    \end{subfigure}
    \hfill
    \begin{subfigure}[b]{0.48\textwidth}
        \centering
        \includegraphics[page=2, width=\textwidth]{images/lianliankan2.pdf}
        \caption{Wrong Action}
        \label{llk:subfig2}
    \end{subfigure}
    \caption{Rules of Shisen-Sho-CIFAR10.}
    \label{fig:llk}
\end{figure}

\paragraph{Rule} In Shisen-Sho, the player clears an 8×8 grid by removing tiles in matching pairs: a pair may be removed only if the two tiles are identical in both shape and color and if there exists a connecting path between them that runs orthogonally (up, down, left, right), makes at most two 90° turns, and passes through no other tiles (see Fig.~\ref{llk:subfig1}).  For instance, two green circles can be cleared when linked around an empty corridor.  If the shapes or colors differ, or if no unobstructed path with $\le2$ turns exists (Fig.~\ref{llk:subfig2}), the move fails and the board remains unchanged. As shown in Fig.~\ref{fig:llk}, the same rules apply in the Shisen-Sho-CIFAR10 variant, except that each tile displays a CIFAR-10 image: a valid match must have the same image class (e.g.\ “cat,” “airplane”). 

\begin{figure}[t]
    \centering
    \begin{subfigure}[b]{0.48\textwidth}
        \centering
        \includegraphics[page=1, width=\textwidth]{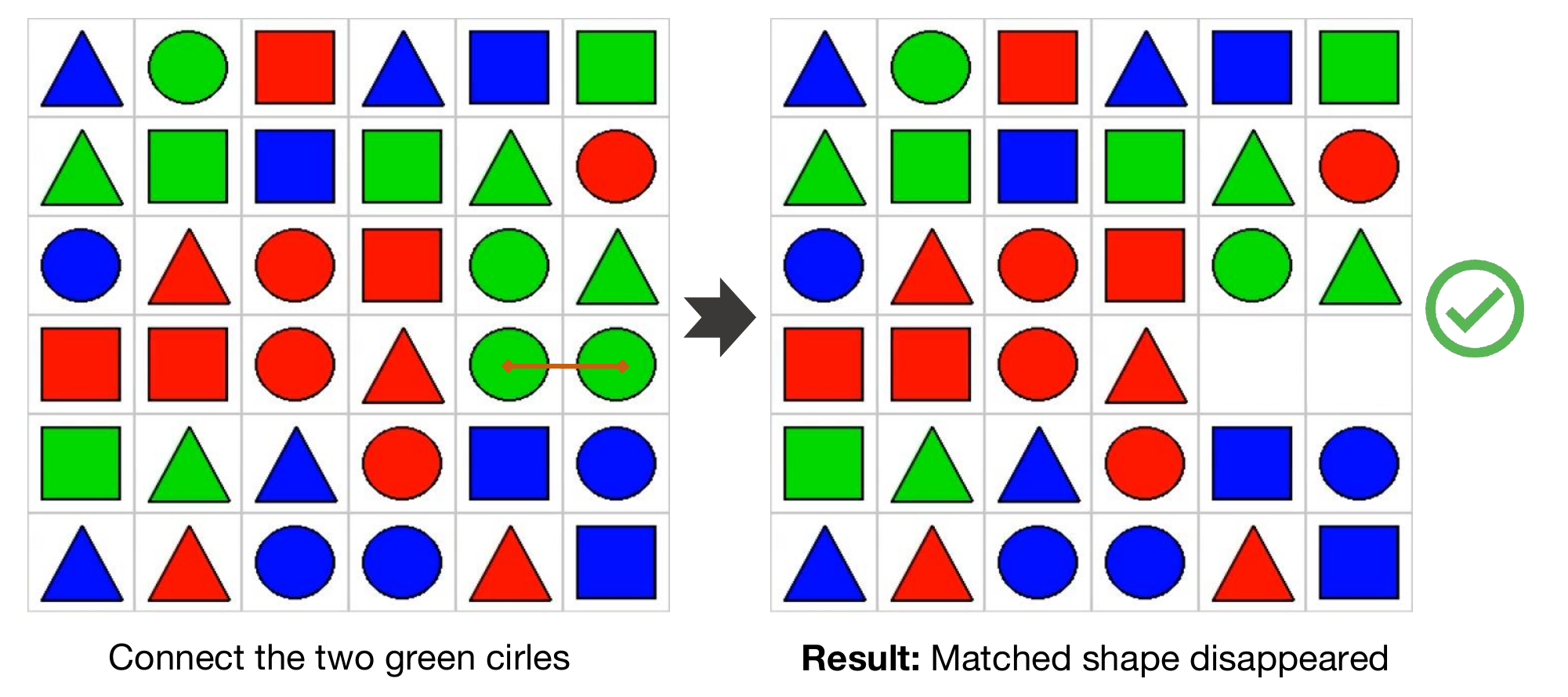}
        \caption{Correct Action}
        \label{llk:subfig1}
    \end{subfigure}
    \hfill
    \begin{subfigure}[b]{0.48\textwidth}
        \centering
        \includegraphics[page=2, width=\textwidth]{images/lianliankan.pdf}
        \caption{Wrong Action}
        \label{llk:subfig2}
    \end{subfigure}
    \caption{Rules of Shisen-Sho.}
    \label{fig:whole}
    \vspace{-5pt}
\end{figure}

\paragraph{Model Play} 
The model receives an 8×8 grid as input, where cells contain colored shapes or CIFAR-10 images. It outputs the coordinates of a matching pair(same shape with same color, or same CIFAR class) that can be connected by an orthogonal path with at most two 90° turns and no occupied cells in between, in the format "(row1,col1) (row2,col2)".

\paragraph{Game Prompt} Here is the game prompt of Shisen-Sho and Shisen-Sho-Cifar10 for models:

\begin{AIbox}{Game Prompt for Shisen-Sho}

\textbf{Game Rule Description}

You are playing a Shisen-sho puzzle game.  The objective is to match pairs of identical tiles by connecting them with a path that has at most 2 turns and doesn't cross any other tiles.

The tiles are distinguished by their color and shape:

- Color include: Red, Green, Blue, Yellow, Magenta, Cyan, etc.

- Shapes include: circle, square, triangle, diamond, cross, star, etc.

Please analyze the game board and identify two matching tiles that can be connected according to these rules.

Return your answer as follows:

1. First coordinate: (row1, col1)

2. Second coordinate: (row2, col2)

Where row and col are 0-indexed numbers such as (0, 1), starting from the top-left of the board. \\

\textbf{Output Format Description}

First describe the board in <perception></perception>. Then output your thinking process in <think></think> and final action in <answer>(row1, col1) (row2, col2)</answer>.

\end{AIbox}

\begin{AIbox}{Game Prompt for Shisen-Sho-Cifar10}

\textbf{Game Rule Description}

You are playing a Shisen-sho puzzle game that uses CIFAR-10 images. Each tile on the board corresponds to one of the CIFAR-10 classes: airplane, automobile, bird, cat, deer, dog, frog, horse, ship, and truck. The objective is to find a pair of tiles that belong to the same class and can be connected with a path that does not cross any other tiles and makes at most two turns.

Please analyze the game board and identify two matching tiles that can be connected according to these rules.

Return your answer as follows:

1. First coordinate: (row1, col1)

2. Second coordinate: (row2, col2)

Where row and col are 0-indexed numbers such as (0, 1), starting from the top-left of the board. \\

\textbf{Output Format Description}

First describe the board in <perception></perception>. Then output your thinking process in <think></think> and final action in <answer>(row1, col1) (row2, col2)</answer>.

\end{AIbox}

\subsection{Swap}
\begin{figure}[t]
    \centering
    \begin{subfigure}[b]{0.48\textwidth}
        \centering
        \includegraphics[page=1, width=\textwidth]{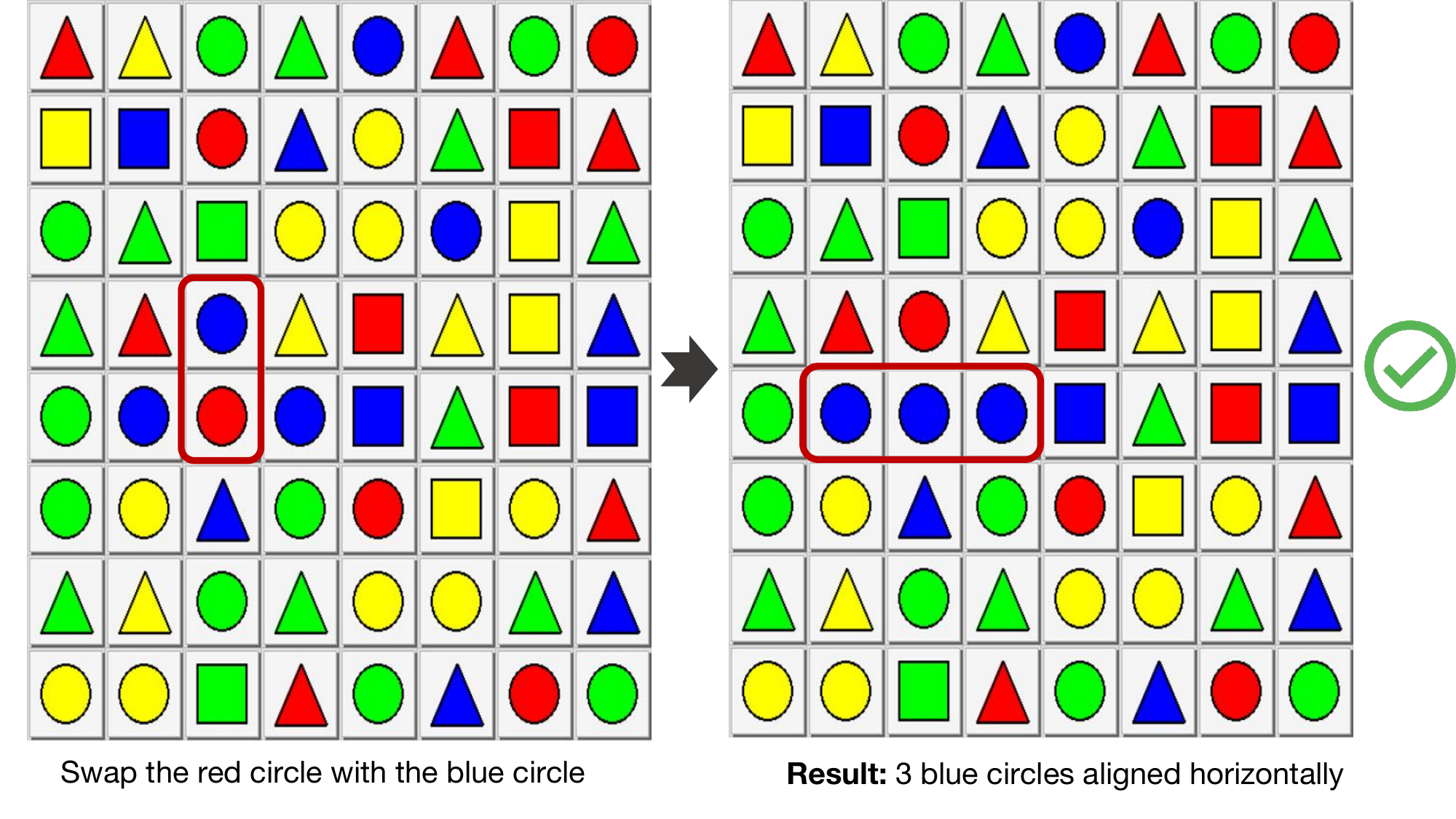}
        \caption{Correct Action}
        \label{fig:subfig1}
    \end{subfigure}
    \hfill
    \begin{subfigure}[b]{0.48\textwidth}
        \centering
        \includegraphics[page=2, width=\textwidth]{images/swap.pdf}
        \caption{Wrong Action}
        \label{fig:subfig2}
    \end{subfigure}
    \caption{Rules of Swap.}
    \label{fig:whole}
\end{figure}

\paragraph{Rule} The player may select any two shapes that share a side (i.e.\ are orthogonally adjacent) and swap them—but only if doing so immediately creates at least one straight line (horizontal or vertical) of three or more identical shapes.  
For example, in Fig.~\ref{fig:subfig1}, swapping the red circle with its adjacent blue circle produces a horizontal row of three blue circles, so the move succeeds. By contrast, in Fig.~\ref{fig:subfig2}, swapping the yellow circle with the adjacent yellow triangle does not form any three‐in‐a‐row; this swap therefore fails, and the two shapes revert to their original positions.

\paragraph{Model Play} 
The model receives an $m$×$n$ board filled with colored shapes as input. It outputs the coordinates of the first valid adjacent tile swap that creates a match-3 or longer, in the format "(row1,col1) (row2,col2)".

\paragraph{Game Prompt} Here is the game prompt of Swap for models:

\begin{AIbox}{Game Prompt for Swap}

\textbf{Game Rule Description}

You are playing a Swap Game where you need to swap adjacent tiles to create matches of 3 or more identical tiles.

- Tiles are identified by color (Red, Green, Blue, Yellow) and shape (circle, square, triangle)

- You can only swap adjacent tiles (not diagonal)

- A valid move must create at least one match of 3 or more identical tiles

- After matches are removed, tiles above will fall down and new tiles will appear at the top

- If no valid moves are available, the board will automatically be shuffled

- The game ends when you run out of moves
        
Please analyze the game board and identify two adjacent tiles to swap that will create a match.

Return your answer as follows:

1. First coordinate: (row1, col1)

2. Second coordinate: (row2, col2)
        
Where row and col are 0-indexed numbers starting from the top-left of the board.
\\

\textbf{Output Format Description}

First describe the board in <perception></perception>. Then output your thinking process in <think></think> and final action in <answer>(row1, col1) (row2, col2)</answer>.

\end{AIbox}

\label{promps_all}
